\IfFileExists{tex/main.tex}{%
  \def\manuscriptdir{tex}%
  \def\bibliographyprefix{tex/}%
  \def\figurepath{figures}%
}{%
  \def\manuscriptdir{.}%
  \def\bibliographyprefix{}%
  \def\figurepath{../figures}%
}
\makeatletter
\edef\input@path{{\manuscriptdir/}}
\makeatother
\documentclass[letterpaper,journal]{IEEEtran}
\usepackage{color}
 \usepackage[final]{changes}
\usepackage{cite}
\usepackage{amsmath,amssymb,amsfonts}
\usepackage{graphicx}
\usepackage{pgf}
\usepackage{placeins}
\usepackage{url}
\usepackage{algorithm}
\usepackage{algpseudocode}
\algrenewcommand\algorithmicrequire{\textbf{Input:}}
\graphicspath{{\figurepath/}{\figurepath/../}}

\newcommand{\T}{\mathbf{T}}
\newcommand{\R}{\mathbf{R}}
\newcommand{\p}{\mathbf{p}}
\newcommand{\q}{\mathbf{q}}
\newcommand{\n}{\mathbf{n}}

\begin{document}
%\title{An Auxiliary Tightening Robot for Modular Lattice Reconfiguration: Mechanism Design, Contact-Constrained Control, and Value-Guided Planning}
\title{CuBot-STT: Simultaneously Transport and Tighten for Lattice-type Modular Reconfiguration Robot}
\author{Chongxi~Meng$^{1,*}$,
    Da~Zhao$^{2,*}$,
    Yanmin~Zhou$^{3}$,
    Runjie~Shen$^{1}$,
    Zhipeng~Wang$^{1,\dagger}$,
    Bin~He$^{1}$,~\IEEEmembership{Senior~Member,~IEEE}%
\thanks{This work was supported in part by the National Natural Science Foundation of China under Grant Nos. 62673365, 62688201, 62403366, 62473294, and 62403363, and in part by the Fundamental Research Funds for the Central Universities.}%
\thanks{$^{*}$Chongxi Meng and Da Zhao contributed equally to this work and are co-first authors.}%
\thanks{$^{1}$Shanghai Research Institute for Intelligent Autonomous Systems, Tongji University, Shanghai, China. $^{2}$Department of Mechanical and Automation Engineering and T Stone Robotics Institute, The Chinese University of Hong Kong, Hong Kong. $^{3}$Department of Control Science and Engineering, Tongji University, Shanghai, China. $^{\dagger}$Corresponding author Zhipeng~Wang: wangzhipeng@tongji.edu.cn}}
\markboth{IEEE/ASME Transactions on Mechatronics}%
{Transport Robot}
\maketitle

\begin{abstract}
Lattice-type modular reconfigurable robots rearrange standardized modules across tasks.
Dedicated transport robots reduce module weight and cost by eliminating module level reconfiguration actuators.
But existing systems use low capacity connections or internal fastening robots that occupy space for functional components.
This paper presents CuBot-STT, which integrates module transport, interface locking, and release in one external robot.
Cubic modules provide structural support and $24\,\mathrm{V}$ power distribution.
On each face, a gear transmission converts one external input into motion of four helical locking points, preserving internal space.
The robot uses alternating foot support, a gripper for module handling, and a tightening head for interface actuation.
The controller coordinates two-foot support and module locking through continuous constraint transfers.
The planner decomposes relocation costs into an analytical lower bound and a residual learned through expert-guided, goal conditioned Q-learning.
Contact face $A^*$ search expands relocation sequences into robot primitives satisfying connectivity, asymmetric manipulation, and foot stance constraints.
Experiments demonstrate normal and tangential interface capacities of $300\,\mathrm{N}$ and $490\,\mathrm{N}$, electrical continuity across six modules, and autonomous reconfiguration of an 11-module lattice.
Planning benchmarks show superior search efficiency over analytical guidance and bidirectional heuristic search, with generalization beyond training lattice sizes without fine-tuning.
\end{abstract}


\section{Introduction}
\label{sec:intro}

Modular reconfigurable robots (MRRs) adapt their morphology to different environments and tasks by rearranging standardized units. This capability supports locomotion, manipulation, and structural construction~\cite{liang2026mrr}. Conventional self-reconfigurable robots typically integrate locomotion actuators, active docking mechanisms, a power supply, and a controller into each module~\cite{yim2007msr,Zhao2024SnailinspiredRS,zhao2022snailbot}. As the number of modules or the required structural load increases, this architecture incurs substantial self-weight and hardware cost, reducing load-bearing efficiency. External assembly architectures~\cite{werfel2014termes,jenett2019,gregg2024armadas} address these limitations by assigning module rearrangement to dedicated robots that move over a lattice of reversibly connected modules~\cite{cheung2013cellular}. The structural units can then provide lightweight support and electrical power distribution. Integrating joints, wheeled drives, and sensing units further enables large-scale multifunctional robotic systems~\cite{smith2024collaborative,meng2026morphogenetic}. The central challenge is to combine module transport with rigid electromechanical interface locking and release in a single external robot. This must be achieved while preserving internal space for functional components.

Existing material--robot systems exploit the relationship between robot motion and module geometry to reduce dependence on global localization. Their interfaces, however, rarely combine high load capacity, electrical continuity, and robotic release. Passive magnetic connections enable inchworm robots to traverse a lattice and transport units but provide limited load capacity~\cite{jenett2019}. Electropermanent magnetic docking within a module provides approximately $88.4\,\mathrm{N}$ of normal holding force, with limited shear capacity~\cite{tosun2016epface}. Snap fits facilitate robotic insertion of new units~\cite{smith2025hierarchical}. Together with separate connectors or reversible thermal soldering, they also allow PCB lattice units to integrate structural, power, and signal functions~\cite{smith2025vik,smith2024collaborative}. However, snap-fit release still requires manual unfastening~\cite{smith2025hierarchical}. These connection approaches also involve trade-offs in load capacity, repeated assembly and disassembly, and energy consumption relative to mechanical fastening. A comparative study of architectural lattice systems likewise shows that the practical benefits of robotic assembly depend on the connection type and system implementation~\cite{smith2026comparative}. An interface that is physically reversible is not necessarily one that a robot can both lock and release. Reconfiguration therefore requires a complete sequence of tool engagement, locking, active release, and module extraction. Functional lattices subject to frequent reconfiguration and substantial loads consequently require rigid electromechanical interfaces that support high-current conduction and robotic locking and release.

Transferring fastening actuators from modules to mobile robots simplifies module design but creates a conflict between fastening access and functional integration. External bipedal inchworm robots can transport and place units on a lattice~\cite{park2023solle}, whereas miniature robots operating inside the lattice actuate captive fasteners~\cite{formoso2023mmici}. Combining these roles, Gregg et al.~\cite{gregg2024armadas} demonstrated construction and reconfiguration of a 256-unit structure using internal robots for fastening and external robots for transport. This division of labor is effective for hollow structural trusses. Functional modules, however, may contain drive motors, batteries, control boards, or specialized payloads, leaving insufficient space for an internal robot and its tools. A single external transport robot (TR) must therefore perform module transport, interface locking, and active release while preserving the module interior for functional payloads.

Combining these functions in one robot introduces coupled execution and planning constraints. During each relocation, physical constraints transfer among the two feet, gripper, tightening tool, and module interfaces. The mechanism must provide external access to the interfaces, and the controller must coordinate module positioning, tool engagement, and locking. At the task level, the relocation sequence~\cite{werfel2014termes,abdelrahman2022swarms} must preserve structural connectivity while satisfying two-foot stance, asymmetric pickup and placement adjacency, and approach-direction constraints~\cite{garcia2025,meng2026morphogenetic}. Weighted single-robot scheduling under connectivity constraints has been shown to be NP-hard~\cite{becker2025tiles}, making large configuration spaces challenging for conventional heuristic search. Moreover, each discrete relocation must be expanded into a collision-free contact-face path and continuous joint trajectories.

\begin{figure}[!t]
\centering
\includegraphics[width=\columnwidth]{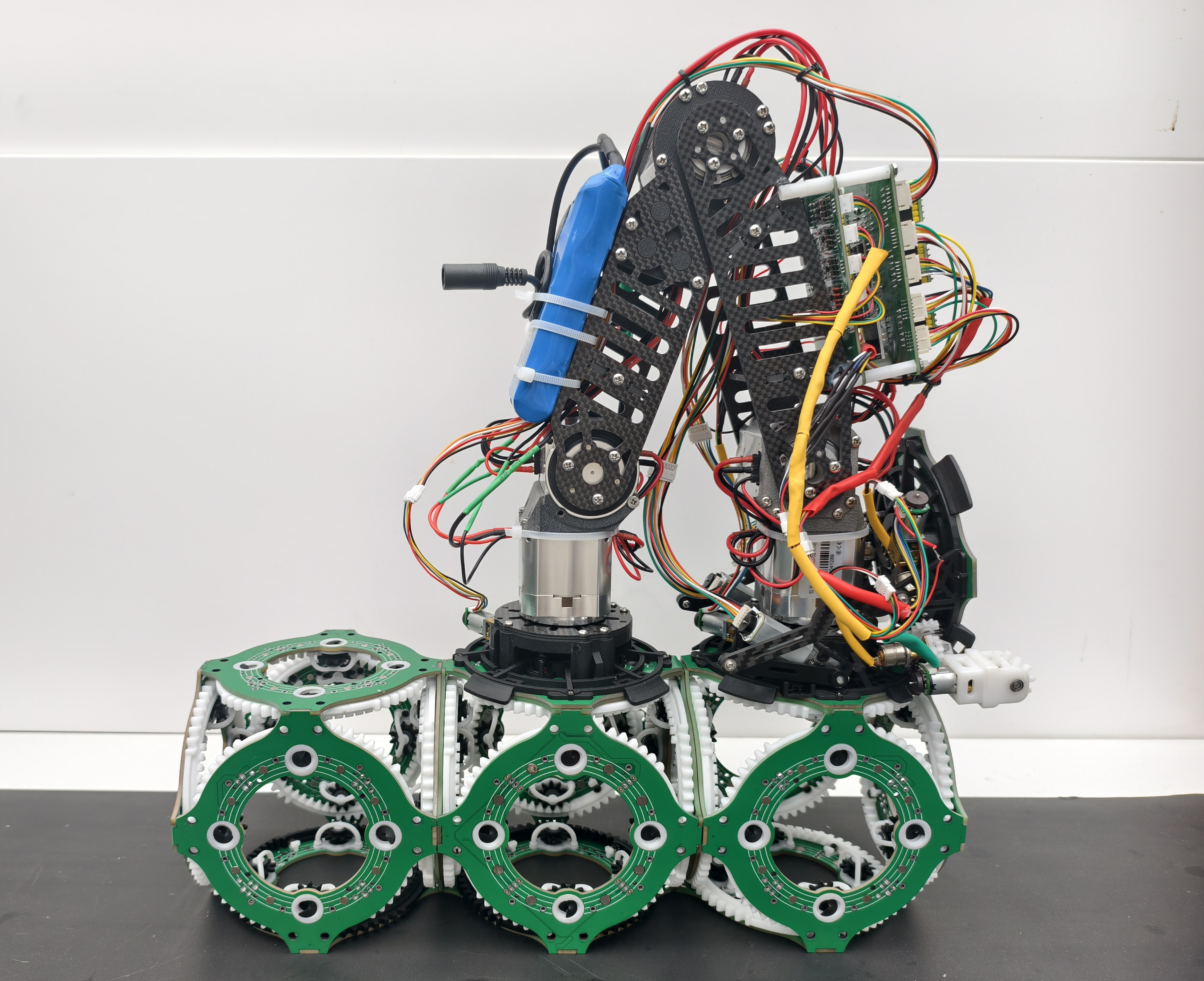}
\caption{CuBot-STT experimental platform. The TR is supported on a lattice of connected CMUs; its locomotion chain and foot interfaces are indicated.}
\label{fig:prototype}
\end{figure}

To address these challenges, this paper presents CuBot-STT, an externally driven lattice system comprising cubic module units (CMUs) and a bipedal alternating-support TR (Fig.~\ref{fig:prototype}). A hierarchical planning and control framework coordinates reconfiguration and execution. The main contributions are as follows:
\begin{enumerate}
  \item \textbf{Externally operated interface--transport robot co-design:} An electromechanical interface and transport robot are co-designed to integrate rigid locking, power transfer, and reversible module unit transport while preserving internal module unit space for functional payloads.
  \item \textbf{Value-guided hierarchical reconfiguration planning:} A hierarchical planner combines expert-guided residual value learning with contact-face motion planning to generate executable relocation sequences under connectivity, asymmetric manipulation, and two-foot support constraints.
  \item \textbf{Prototype and large-scale validation:} A physical prototype and large-scale benchmarks validate interface strength, electrical continuity, autonomous reconfiguration, and planning scalability, demonstrating the feasibility of externally driven assembly of functional modular lattices.
\end{enumerate}

The remainder of this paper is organized as follows. Section~\ref{sec:mechanism} introduces the module units, transport robot mechanical design, and operation sequence. Section~\ref{sec:system-control} describes the hierarchical control architecture, contact-constrained kinematics, expert-guided value-learning configuration search, and contact-face atomic action expansion. Section~\ref{sec:experiments} reports interface mechanical and electrical tests, physical transport demonstrations, and planning algorithm evaluations. Section~\ref{sec:conclusion} concludes the paper and discusses future work.

\section{Mechanical Design}
\label{sec:mechanism}

\begin{table}[!t]
\renewcommand{\arraystretch}{1.15}
\caption{Main Mechanical Design Parameters}
\label{tab:module-params}
\centering
\footnotesize
\setlength{\tabcolsep}{3.5pt}
\begin{tabular}{@{}llp{0.28\columnwidth}@{}}
\hline
Symbol & Meaning & Value \\
\hline
\multicolumn{3}{@{}l}{Module unit} \\
$d$ & Module unit side length / lattice pitch & $120\,\mathrm{mm}$ \\
$z_o$ & External ring gear teeth & $59$ \\
$z_r$ & Internal ring gear teeth & $59$ \\
$n_h$ & Number of helical pinions & $4$ \\
$z_h$ & Helical pinion teeth & $10$ \\
$|\theta_g|$ & Main gear locking angle & $40^\circ$ \\
$|\theta_h|$ & Helical pinion locking angle & $236^\circ$ \\
$\delta_h$ & Locking extension & $3.05\,\mathrm{mm}$ \\
\hline
\multicolumn{3}{@{}l}{Locomotion, pickup and placement, and interface actuation} \\
$L$ & Main leg length & $170\,\mathrm{mm}$ \\
$\alpha_{\max}$ & Four-bar flipping angle & $97^\circ$ \\
$s_{\max}$ & Timing-belt clamping stroke & $14.69\,\mathrm{mm}$ \\
$z_p$ & Output pinion teeth & $10$ \\
$\delta_e$ & Tightening head engagement stroke & $5\,\mathrm{mm}$ \\
$|\theta_m|$ & Locking motor shaft angle & $236^\circ$ \\
\hline
\end{tabular}
\end{table}

\begin{figure*}[!t]
\centering
\includegraphics[width=\textwidth]{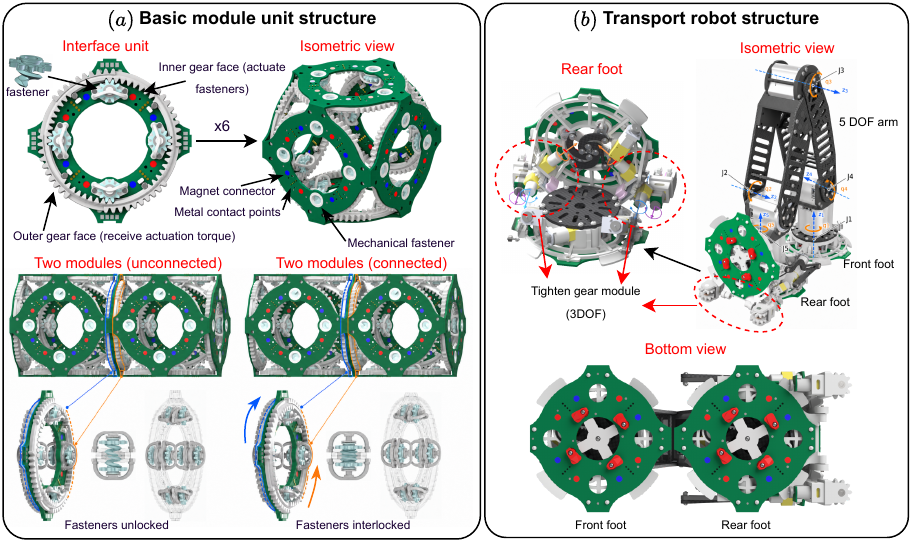}
\caption{Mechanical design of the CMU, electromechanical interface, and TR. (a) CMU face, assembled cube, interface locking states, and transmission details. Curved arrows indicate the tightening directions. (b) TR architecture and foot, gripping, and tightening mechanisms.}
\label{fig:structure}
\label{fig:module-unit-array}
\label{fig:interface}
\end{figure*}

\begin{figure*}[!t]
\centering
\includegraphics[width=\textwidth]{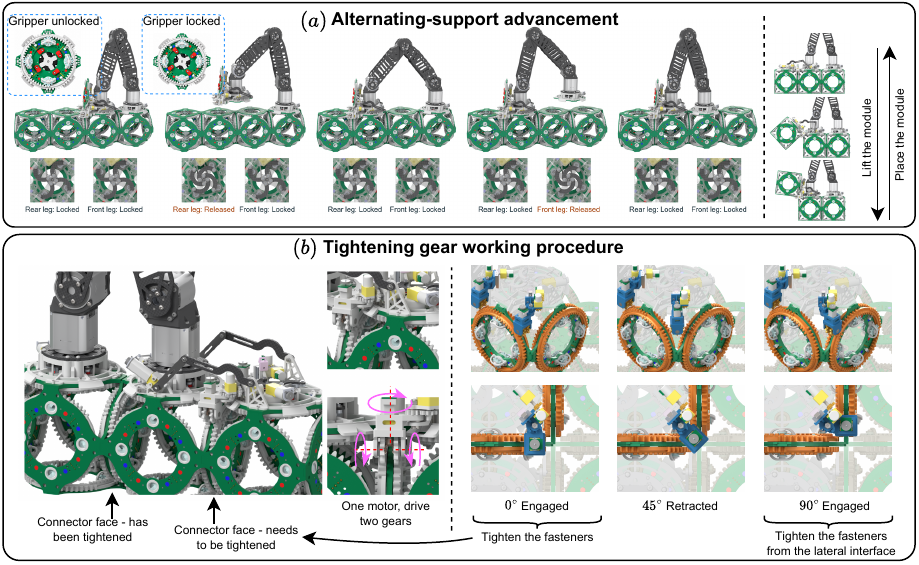}
\caption{CMU handling and alternating-support locomotion. (a) CMU lifting, five locomotion states, and released and clamped foot configurations; $s_{\max}$ denotes the timing-belt clamping stroke. (b) Pickup and placement poses and tightening-head engagement states at $0^\circ$, $45^\circ$, and $90^\circ$.}
\label{fig:lift}
\label{fig:working}
\label{fig:toolsequence}
\label{fig:step}
\label{fig:footlock}
\end{figure*}

\subsection{Cubic Module Unit Design}

CMUs are arranged on a lattice with pitch $d$ and use a common interface design on all faces. Each CMU consists of six structurally identical PCB panels assembled into a cube (Fig.~\ref{fig:module-unit-array}(a)). The panels provide both structural support and electrical conduction; Table~\ref{tab:module-params} lists the principal dimensions and actuation parameters. Tabs along adjacent panel edges meet at right angles and are soldered together. The solder joints along the twelve cube edges maintain the geometry and electrically connect the faces, eliminating separate internal wiring.

Each CMU face carries a double-sided locking transmission on its inner surface. The main gear is coaxial with the panel and has concentric external and internal toothed rings. An external tightening head drives the external ring, while the internal ring drives $n_h$ uniformly distributed helical pinions. A main-gear rotation of $\theta_g$ produces a pinion rotation of $|\theta_h|=(z_r/z_h)|\theta_g|$, where $z_r$ and $z_h$ denote the tooth counts of the internal ring and helical pinion, respectively. A single external input thus extends or retracts the four locking points by $\delta_h$. Neither tool access to the CMU interior nor a motor on each face is required.

Magnets provide passive alignment before docking, and spring-loaded pins establish electrical contact. All faces share a layout with $90^\circ$ rotational symmetry, with opposite magnetic poles facing each other in all four docking orientations. The interfaces therefore require no distinction between male and female faces. After magnetic attraction provides preliminary positioning, the spring-loaded pins compress into contact and the helical transmission locks the interface mechanically. Current passes through the pins to the neighboring CMU and reaches the other faces through the soldered edge connections.

\subsection{TR Mechanisms and Operation Sequences}
\label{sec:robot}
\label{sec:operation}

The TR integrates lattice locomotion, CMU handling, and interface actuation on one external platform (Fig.~\ref{fig:structure}(b)). Its front and rear support feet are denoted by $F$ and $B$, respectively, and its vertical gripper by $E$. The feet anchor the TR to the lattice, while the gripper holds and manipulates the target CMU.

The locomotion chain connects the two feet through five revolute joints, $J_1$--$J_5$, driven by planetary geared motors. The separate pickup and placement joint, $J_6$, maintains the gripper orientation during stepping. Each foot clamps a CMU face using a Hall-sensor geared motor, timing belt, turntable, and locking claws (Fig.~\ref{fig:structure}(b)). These clamps anchor the TR to the contacted CMU; support through the lattice also requires the neighboring CMU interfaces to remain locked.

A four-bar linkage driven by two Hall-sensor geared motors performs CMU handling. A separate dual-gear tightening head, driven by symmetric motors through three-stage transmissions, engages the external ring gears on both sides of a docking seam. The head operates entirely from outside the structure. Through the output pinion ($z_p$) and external ring gear ($z_o$), a $236^\circ$ motor rotation produces a $40^\circ$ main-gear rotation and completes the locking stroke (Table~\ref{tab:module-params}).

Operation is organized into locomotion and pickup and placement primitives (Fig.~\ref{fig:lift}). Locomotion changes the foot contact faces: same-face translation advances the TR by one lattice cell (Fig.~\ref{fig:step}(a)), while same-face yaw changes its in-plane travel direction. During pickup and placement, both feet remain clamped. The four-bar linkage and gripper transfer the target CMU between support by its interfaces and support by the gripper. In each operation, the next supporting or gripping constraint is established before the preceding constraint is released. This sequence keeps the TR supported and the handled CMU constrained.

A step begins with the rear foot clamped while the front foot swings to the target contact face and clamps. Support then transfers to the front foot, allowing the rear foot to advance and reclamp. The cycle ends with both feet clamped (Fig.~\ref{fig:step}(a)).

To release and remove a CMU, the four-bar linkage positions the gripper, which clamps and secures the unit. The tightening head then engages and releases the inter-CMU interface before retracting. The linkage subsequently lifts the CMU. During placement, the linkage lowers the CMU to the target position, the tightening head locks the interfaces, and the gripper releases the unit (Fig.~\ref{fig:working}(b)). The head operates at the two orthogonal orientations of $0^\circ$ and $90^\circ$. When a CMU has multiple contact faces, the head actuates the interfaces sequentially.

\section{System Control and Reconfiguration Planning}
\label{sec:system-control}
\label{sec:planning}

\subsection{Control Architecture and Contact-Constrained Kinematics}
\label{sec:control}
\label{sec:ik}
The control system adopts centralized planning and hierarchical execution. The planning layer generates point-to-point macro-actions (Section~\ref{sec:value-guided-search}) and expands them into contact-face atomic actions (Section~\ref{sec:astar}), which the host computer dispatches one at a time on completion events. The kinematic layer precomputes continuous joint trajectories for two execution boards that track them with local time bases, while the foot clamps, gripper, and tightening head execute stage commands.

Locomotion and pickup and placement are solved under contact constraints: one foot is fixed as the base, and the other swings toward the target CMU face. To match the five-DoF kinematics of the locomotion chain with the discrete symmetry of the CMU faces, each foot contact target is parameterized by its face center, outward normal, and aligned in-plane edge direction, and continuous trajectories are generated subject to joint limits and collision avoidance. The locomotion variables are $\q=(q_1,\ldots,q_5)^\mathsf{T}$; the pickup and placement joint $q_6$ is excluded.

Let $W$ denote the world frame and $R$ the model root frame, $\T_{Rf}(\q)$ be the forward kinematics of foot $f$ relative to the root, and $\bar{\T}_{W\mathrm{fix}}$ the anchored pose of the current fixed foot $a_{\mathrm{fix}}$. After updating the locomotion joints, the root is reconstructed using
\begin{equation}
 \T_{WR}(\q)=\bar{\T}_{W\mathrm{fix}}\T_{R\mathrm{fix}}(\q)^{-1}
\label{eq:anchor}
\end{equation}
to keep the position and orientation of the fixed foot unchanged during the swing. When the front foot is fixed, calculation proceeds from the front foot to the rear foot; when the rear foot is fixed, the same joint transformations are used in reverse order. At handover, only the fixed-foot indicator and anchored pose are updated, yielding the same root transformation through~\eqref{eq:anchor}.

The target contact face is specified by the CMU coordinates $\mathbf{v}=(i,j,k)$ and outward normal $\n$. For lattice pitch $d$, the CMU center and face center are
\begin{equation}
 \p_c=d(i,j,k+\tfrac12)^\mathsf{T},\qquad
 \p_d=\p_c+\frac d2\n.
\label{eq:face}
\end{equation}
The sole target also requires an in-plane orientation: an in-plane unit tangent $\mathbf{t}$ is selected, and the contact-face basis is constructed from $\mathbf{t}$, $\n\times\mathbf{t}$, and $\n$ to obtain the target rotation $\R_d$. Endpoint solving aligns the face center, sole normal, and in-plane edge direction in sequence, and determines the kinematic reachability of the target contact face using joint limits and collision checks.

The position error is $\mathbf{e}_p=\p(\q)-\p_d$, and the orientation error is $\mathbf{e}_R=\operatorname{Log}(\R(\q)\R_d^\mathsf{T})^\vee$. The task errors are assembled into $\mathbf{e}$, and differential inverse kinematics under the fixed-foot condition computes the joint increment for the free foot:
\begin{equation}
\begin{aligned}
 \min_{\Delta\q}\;&\tfrac12\|\mathbf{J}\Delta\q+\gamma\mathbf{e}\|_{\mathbf{W}}^2
                  +\tfrac\lambda2\|\Delta\q\|^2,\\
 \text{s.t.}\;&\mathbf{b}_{\min}\leq\Delta\q\leq\mathbf{b}_{\max},\\
             &\mathbf{A}\Delta\q\leq\mathbf{b}
\end{aligned}
\label{eq:ik}
\end{equation}
where $\mathbf{J}$ is the corresponding Jacobian. Each iteration is followed by a geometric collision check, in which a carried CMU is treated as rigidly attached to the gripper, and a failed check rejects the increment. A step in Fig.~\ref{fig:step}(a) is accepted when both foot transfers obtained from~\eqref{eq:face}--\eqref{eq:ik} maintain alternating support, respect the joint limits, and are collision-free.

\subsection{Expert-Guided Value Learning and Configuration Search}
\label{sec:bidirectional}
\label{sec:value-guided-search}
Configuration planning converts differences between the current and target CMU arrangements into an ordered sequence of point-to-point pickup and placement operations. Each search node is a complete configuration $C$ recording cell occupancy and CMU types, with the target configuration denoted by $G=C_{\mathrm{goal}}$. A macro-action $a=A_H=(\mathbf{p}_{\mathrm{pick}},\mathbf{p}_{\mathrm{drop}},\tau)$ transports a CMU of type $\tau$ from the pickup cell to the placement cell, yielding the successor configuration $C'=T(C,a)$; $\mathcal A(C)$ denotes the legal action set. Under four-neighborhood adjacency in the CMU plane, $a\in\mathcal A(C)$ if (i) the picked CMU has at most three neighbors and is not an articulation point, so the remaining structure stays four-connected; (ii) the placement cell has one or two neighbors in the remaining structure; and (iii) an adjacent CMU pair that supports both feet exists beside the pickup cell and, after removal, beside the placement cell.

\textbf{1) Residual action-value model:} With a unit cost for each CMU relocation, the undiscounted goal-conditioned action value $Q_\theta(C,a,G)$ represents the accumulated relocation cost of executing $a$ from $C$ and eventually reaching $G$. With the count of each CMU type fixed, let $M(C',G)$ denote the number of CMUs in the successor configuration whose positions or types do not match the goal. Each such CMU must be relocated at least once, while each macro-action moves only one CMU; therefore, $M$ gives an analytical lower bound on the remaining number of relocations. The action value is decomposed into the current single-step cost, this lower bound, and a nonnegative residual. At the goal configuration, $Q_\theta(G,a,G)=0$; otherwise,
\begin{equation}
\begin{aligned}
 Q_\theta(C,a,G)&=1+M(C',G)+r_\theta(C,a,G),\\
 r_\theta(C,a,G)
 &=\mathbf 1_{\{M(C',G)>0\}}
 \operatorname{softplus}\!\bigl(f_\theta(C,a,G)\bigr).
\end{aligned}
\label{eq:residual-q-model}
\end{equation}
Here, $\mathbf 1$ is the indicator function and $\operatorname{softplus}(z)=\log(1+\exp z)$. The residual $r_\theta$ estimates the additional detour cost induced by connectivity and stance constraints; it is disabled when $M=0$, making the cost of an action directly reaching the goal exactly $1$. A shared residual convolutional encoder extracts current--goal spatial features, while a pickup--placement pairing head combines local action features and global context to predict the residual cost.

\textbf{2) Expert-guided joint learning:} The bootstrap target is
\begin{equation*}
 y_{\bar\theta}=
 \begin{cases}
  1, & C'=G,\\
  1+\min_{a'\in\mathcal A(C')}Q_{\bar\theta}(C',a',G),
     & \text{otherwise}.
 \end{cases}
\end{equation*}
The temporal-difference (TD) loss is
\begin{equation}
 \mathcal L_{\mathrm Q}(\theta)
 =\mathbb E_{\mathcal B}
 \!\left[\bigl(Q_\theta(C,a,G)-y_{\bar\theta}\bigr)^2\right],
\label{eq:goal-q-learning}
\end{equation}
where $\theta$ and $\bar\theta$ are the online and target network parameters, respectively, and $\mathcal B$ is the replay set of $(C,a,G)$ samples. Bootstrap updates are performed only when $\mathcal A(C')\neq\emptyset$, preventing topological deadlocks without legal successors from being mistaken for zero-cost terminal states. Teacher trajectories are added to the expert set after legality replay, and for each state--goal pair $(C,G)$, the actions appearing in valid trajectories form a nonempty positive set $\mathcal P(C,G)\subseteq\mathcal A(C)$. To allow multiple known feasible actions to share probability mass, the expert loss maximizes the total selection probability of this set:
\begin{equation}
\begin{aligned}
 \mathcal L_{\mathrm E}(\theta)
 &=-\mathbb E_{\mathcal D_{\mathrm E}}
 \!\left[\log\frac{Z_{\mathrm P}(C,G)}{Z_{\mathrm A}(C,G)}\right],\\
 Z_{\mathrm X}(C,G)
 &=\sum_{a\in\mathcal X}
 \exp\!\bigl(-Q_\theta(C,a,G)/\vartheta\bigr),
\end{aligned}
\label{eq:expert-action-guidance}
\end{equation}
where $\mathcal D_{\mathrm E}$ is the set of expert state--goal pairs, $\vartheta$ is the Boltzmann temperature, and $Z_{\mathrm P}$ and $Z_{\mathrm A}$ sum over $\mathcal P(C,G)$ and $\mathcal A(C)$, respectively. Valid expert trajectories also provide the shortest known feasible length $U(C,a,G)$ from action $a$ to the goal, including the current action, which upper-bounds the optimal action cost. The joint objective is
\begin{equation}
\begin{aligned}
 \mathcal L(\theta)
 &=\mathcal L_{\mathrm Q}(\theta)
 +\lambda_{\mathrm E}\mathcal L_{\mathrm E}(\theta)\\
 &\quad+\lambda_{\mathrm U}\mathcal L_{\mathrm U}(\theta),\\
 \mathcal L_{\mathrm U}(\theta)
 &=\mathbb E_{\mathcal D_{\mathrm E}}
 \!\left[\frac{1}{|\mathcal P(C,G)|}
 \sum_{a\in\mathcal P(C,G)}
 \delta_{\mathrm U}(C,a,G)^2\right],
\end{aligned}
\label{eq:joint-q-objective}
\end{equation}
where $\delta_{\mathrm U}(C,a,G)=\bigl[Q_\theta(C,a,G)-U(C,a,G)\bigr]_+$ and $[x]_+=\max(x,0)$. This one-sided term penalizes only overestimation of expert action costs and does not force the network to fit potentially suboptimal demonstration lengths.

\begin{algorithm}[!t]
\caption{Expert-Guided Residual Value Learning}
\label{alg:value-learning}
\begin{algorithmic}[1]
\State Initialize $\theta$, $\bar\theta\gets\theta$, $\mathcal B$, $\mathcal D_{\mathrm E}$, and curriculum level $k$
\Repeat
 \State Sample $(C_{\mathrm{ini}},G)$ at curriculum level $k$
 \State $\pi\gets$ Algorithm~\ref{alg:value-search} with $Q_\theta$ and a short budget
 \If{$\pi$ fails}
  \State $\pi\gets$ expert search with a long budget
 \EndIf
 \If{$\pi$ is found}
  \State Replay $\pi$ for legality and trim redundant moves
  \State Update $\mathcal B$, $\mathcal P(C,G)$, and $U(C,a,G)$ from $\pi$
 \EndIf
 \State Augment $\mathcal B$ by symmetry and hindsight relabeling
 \State Compute $y_{\bar\theta}$ and update $\theta$ by minimizing~\eqref{eq:joint-q-objective}
 \State Periodically set $\bar\theta\gets\theta$ and adjust $k$ by validation
\Until{training converges}
\end{algorithmic}
\end{algorithm}

\begin{algorithm}[!t]
\caption{Residual-Value-Guided Configuration Search}
\label{alg:value-search}
\begin{algorithmic}[1]
\Require $C_{\mathrm{ini}}$, $G$, frozen $Q_\theta$, time and expansion budgets
\State $g(\cdot)\gets\infty$, $g(C_{\mathrm{ini}})\gets0$; push $(C_{\mathrm{ini}},a)$ for all $a\in\mathcal A(C_{\mathrm{ini}})$
\While{the queue is nonempty and within budget}
 \State Pop $(C,a)$ with minimum $F(C,a)$
 \State $C'\gets T(C,a)$, $g'\gets g(C)+1$
 \If{$g'\geq g(C')$}
  \State \textbf{continue}
 \EndIf
 \State $g(C')\gets g'$ and record $(C,a)$ as the parent of $C'$
 \If{$C'=G$}
  \State \Return backtracked and verified macro-actions
 \EndIf
 \State Evaluate $Q_\theta(C',a',G)$ for all $a'\in\mathcal A(C')$
 \State Push each $(C',a')$ with $F(C',a')=g'+Q_\theta(C',a',G)$
\EndWhile
\State \Return failure
\end{algorithmic}
\end{algorithm}

\textbf{3) Training and search:} Algorithm~\ref{alg:value-learning} summarizes offline curriculum training. The expert planner adapts the bidirectional weighted search in~\cite{meng2026morphogenetic}, reversing pickup and placement neighbor bounds during backward expansion. Algorithm~\ref{alg:value-search} ranks candidate edges by combining the minimum known relocation count $g(C)$ from $C_{\mathrm{ini}}$ to $C$ with $Q_\theta$, which includes the current relocation cost. The resulting macro-actions are expanded into contact-face primitives (Section~\ref{sec:astar}).
\subsection{Atomic Action Expansion}
\label{sec:astar}
To connect discrete macro-actions with low-level continuous control, the mid-level planner expands each point-to-point transport task on a contact-face state graph into a sequence of foot contact-face transfers and pickup and placement primitives.

Each foot corresponds to a contact face $f=(v,\n)$, where $v\in\mathbb Z^3$ is the CMU lattice point and the outward normal is $\n\in\{\pm\mathbf e_x,\pm\mathbf e_y,\pm\mathbf e_z\}$. The TR stance is defined as
\begin{equation}
 s=(f_F,f_B,a_{\mathrm{fix}},\hat u,c),
\label{eq:stance}
\end{equation}
where $f_F$ and $f_B$ are the front and rear foot contact faces, $a_{\mathrm{fix}}\in\{F,B\}$ is the current fixed foot, $\hat u\perp\n$ is the in-plane travel direction, and $c\in\{0,1\}$ indicates whether a CMU is carried. Each macro-action is expanded into an ordered combination of a walking contact-face sequence $\sigma_{\mathrm{walk}}$ and pickup and placement actions $A_{\mathrm{pick/place}}$.

Stepping comprises the complete cycle in Fig.~\ref{fig:step}(a), from the initial two-foot stance through both foot transfers to restored two-foot support. Turning likewise covers a complete cycle ending with both feet clamped at the target orientation. Face unlocking and locking each comprise one complete release or tightening operation on a single CMU interface; their costs add across multiple contact faces. Pickup and placement comprise four-bar motion and gripper clamping and locking or release, with interface actuation counted separately. These operations follow the constraint-transfer sequence in Section~\ref{sec:operation}.

During placement, a jam triggers encoder-based homing before the next attempt. A placement trial is counted as failed after three unsuccessful attempts.

The $A^*$ search proceeds in two stages: it first travels from the current stance to a contact stance allowing pickup and appends the pickup primitive; it then reaches a contact stance allowing placement on the structure after removal of that CMU and appends the placement primitive. The node evaluation and heuristic are
\begin{equation}
\begin{aligned}
 J(s)&=g_{\mathrm{walk}}(s)+h_{\mathrm{walk}}(s),\\
 g_{\mathrm{walk}}(s)&=\sum_{o\in\mathcal O(s)}\hat c(o),\\
 h_{\mathrm{walk}}(s)&=\frac{\hat c_{\min}}{2}
 \bigl(\lvert x_F-x_F^\star\rvert+\lvert y_F-y_F^\star\rvert\\
 &\qquad\quad+\lvert x_B-x_B^\star\rvert+\lvert y_B-y_B^\star\rvert\bigr),
\end{aligned}
\label{eq:walk-astar}
\end{equation}
where $\mathcal O(s)$ is the sequence of primitives executed to reach stance $s$, and $\hat c(o)$ is the time cost of primitive $o$, calibrated using the measured durations in Section~\ref{sec:primitive-eval}. The term in parentheses is the sum of planar Manhattan distances from the front and rear foot contact voxels to their respective target positions; superscript $\star$ denotes the target contact stance of the current task. Each locomotion primitive reduces this distance sum by at most $2$ and has a cost no lower than the minimum locomotion primitive cost $\hat c_{\min}$. Thus, $h_{\mathrm{walk}}$ is measured in seconds and does not overestimate the remaining walking time, satisfying $A^*$ admissibility.

% Keep the four interface validation figures together in a two-by-two grid.
\begin{figure*}[!t]
\centering
\begin{minipage}{\textwidth}
\begin{minipage}[t]{0.48\textwidth}
\centering
\begin{minipage}[c][0.32\textheight][c]{\linewidth}\centering
\includegraphics[width=\linewidth,height=0.32\textheight,keepaspectratio]{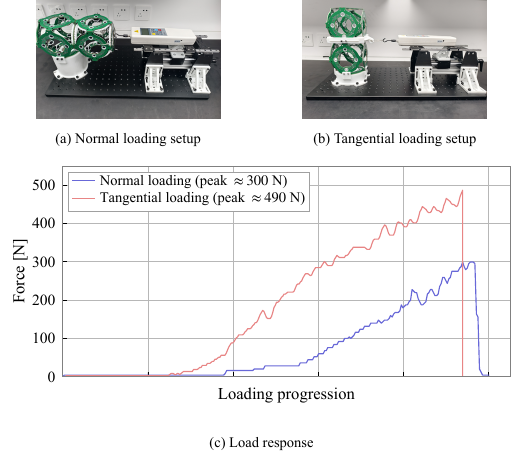}
\end{minipage}
\caption{Ultimate loading tests of the CMU interface. (a) Normal and (b) tangential loading setups. (c) Load--displacement curves.}
\label{fig:interface-force}
\end{minipage}\hfill
\begin{minipage}[t]{0.48\textwidth}
\centering
\begin{minipage}{\linewidth}\centering
\resizebox{\linewidth}{!}{\input{\figurepath/strength_comparison_300N.tex}}
\end{minipage}
\caption{Strength versus density comparison of programmable materials and modular self-reconfigurable robots. The left blue dotted ellipse groups programmable materials, while the right green dotted ellipse groups modular self-reconfigurable robots. The star denotes the proposed CMU; diagonal dashed lines indicate equal specific strength.}
\label{fig:strength-density-comparison}
\end{minipage}
\par\vspace{10pt}
\begin{minipage}[t]{0.48\textwidth}
\centering
\begin{minipage}[c][0.27\textheight][c]{\linewidth}\centering
\includegraphics[width=\linewidth,height=0.27\textheight,keepaspectratio]{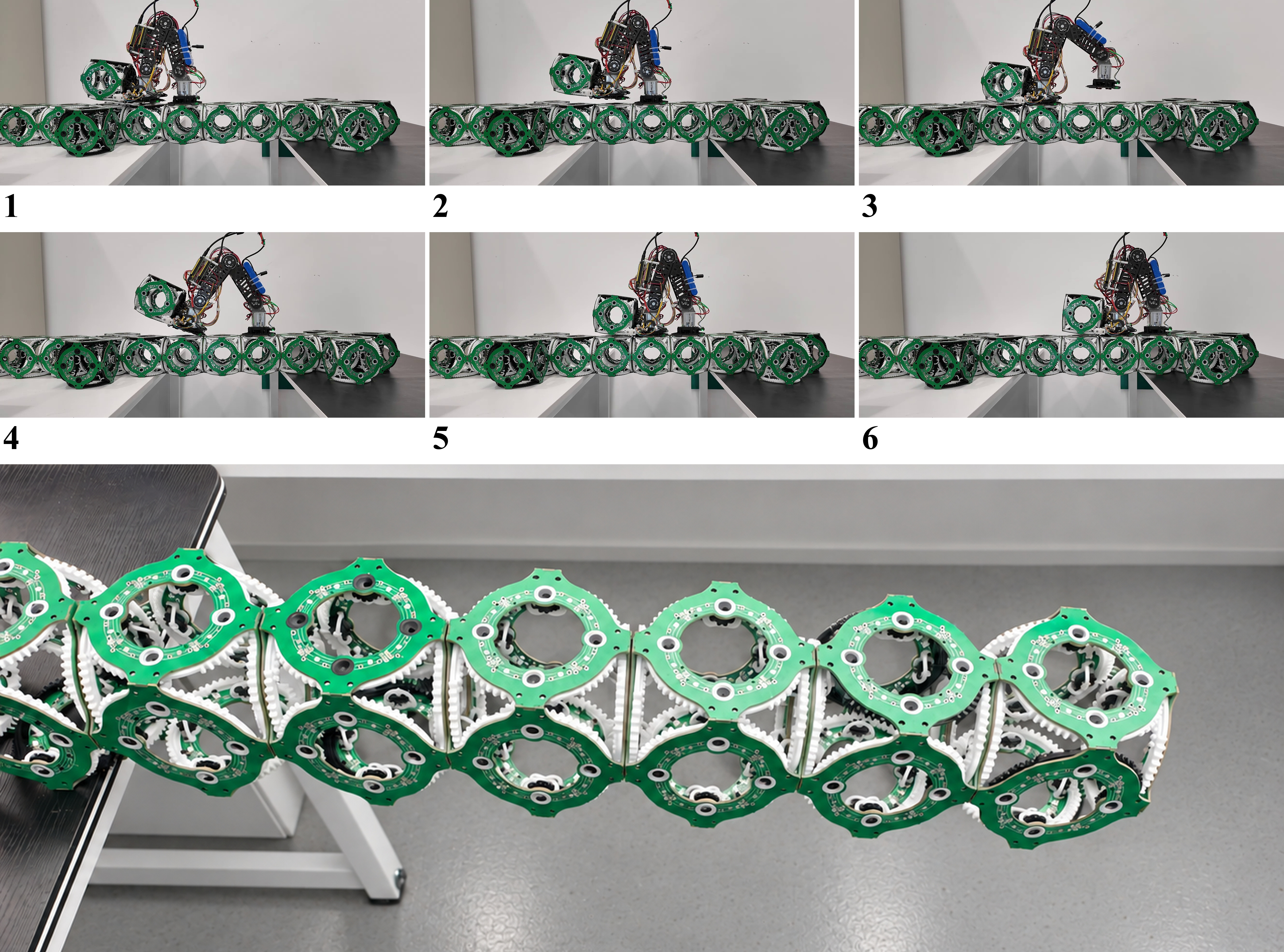}
\end{minipage}
\caption{Application-level validation of interface strength: moving-load support on a beam spanning two tables (frames 1--6) and a cantilever supporting six suspended CMUs.}
\label{fig:bridge-load}
\end{minipage}\hfill
\begin{minipage}[t]{0.48\textwidth}
\centering
\begin{minipage}[c][0.27\textheight][c]{\linewidth}\centering
\includegraphics[width=\linewidth,height=0.27\textheight,keepaspectratio]{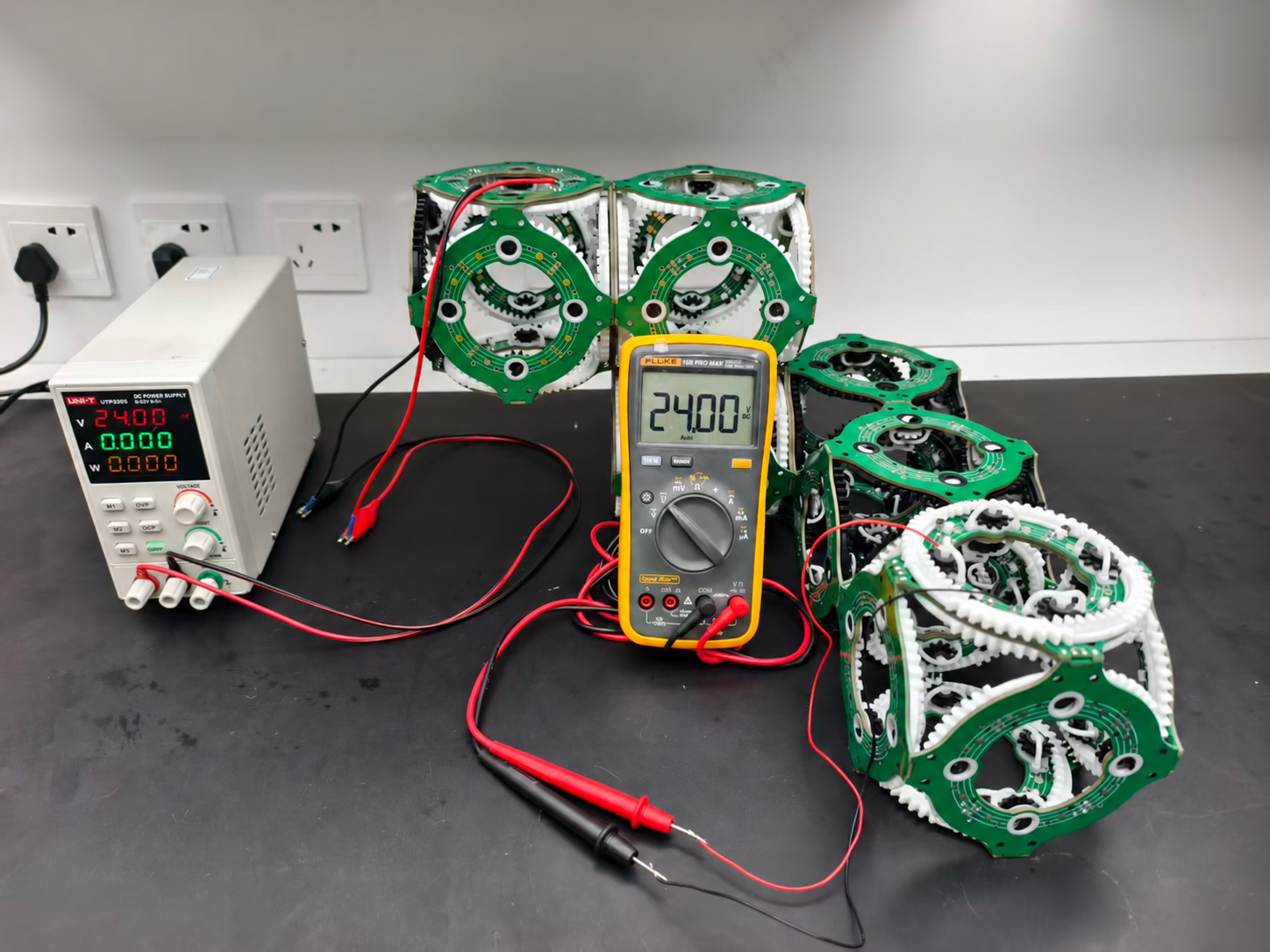}
\end{minipage}
\caption{Electrical continuity test across a six-module-unit cantilever with a $24\,\mathrm{V}$ supply applied at the free end.}
\label{fig:power-photo}
\end{minipage}
\end{minipage}
\end{figure*}

\section{Experimental Results}
\label{sec:experiments}

\subsection{Structural Load Capacity and Electrical Continuity Across Interfaces}
\label{sec:interface-eval}

To evaluate the mechanical strength of the CMU interface, the base CMU was fixed to a test bench, and a force gauge and loading platform applied normal tensile and tangential offset loads (including both shear and bending moments) to an adjacent connected CMU, as shown in Fig.~\ref{fig:interface-force}(a), (b). The load--displacement response curves (Fig.~\ref{fig:interface-force}(c)) show ultimate normal and tangential capacities of $300\,\mathrm{N}$ and $490\,\mathrm{N}$, respectively. Both failure modes involved structural fracture of the interface helical pinion at its central shaft connection; the ultimate interface capacity is therefore limited by the structural strength of the helical pinion body.

Fig.~\ref{fig:bridge-load} validates interface strength in practical applications: the TR carries a CMU along a beam spanning two tables, and a cantilever supports six suspended CMUs. The connections remain stable without slippage or separation.

\paragraph{Strength--Density Comparison}
To examine the tradeoff between structural load capacity and lightweight design, Fig.~\ref{fig:strength-density-comparison} summarizes the strength--density distribution of reconfigurable systems following the comparison method of Gregg et al.~\cite{gregg2024armadas}. The CMU has a side length of $120\,\mathrm{mm}$, a mass of $280\,\mathrm{g}$, and a structural envelope density of $\rho=280/12^3=0.162\,\mathrm{g/cm^3}$. Normalizing the $300\,\mathrm{N}$ peak tensile capacity by the CMU face area ($d^2$) yields a nominal interface tensile strength of $\sigma_{\mathrm{nom}}=F_n/d^2\approx20.8\,\mathrm{kPa}$ (star in Fig.~\ref{fig:strength-density-comparison}), corresponding to a specific strength of $\sigma_{\mathrm{nom}}/\rho\approx0.129\,\mathrm{kN\,m/kg}$.

The diagonal dashed lines indicate equal specific strength $\sigma/\rho$; movement toward the upper left represents higher nominal strength at lower density.

Mechanically locked CMUs also establish electrical bus connections. In the six-CMU cantilever test (Fig.~\ref{fig:power-photo}), a DC supply was connected to the suspended end CMU (output set to $24.0\,\mathrm{V}$). After cascaded transmission through PCB metal contacts between CMUs, the measured voltage at the base end was $24\,\mathrm{V}$, with negligible voltage drop.

\subsection{Primitive Reliability, Cost Calibration, and Payload Response}
\label{sec:primitive-eval}

With structural load and power transmission across interfaces verified in Section~\ref{sec:interface-eval}, this subsection quantifies the action primitives that lock, release, and carry CMUs on the lattice; the measured durations also provide the edge costs used by the mid-level planner.

\begin{table}[!t]
\centering
\caption{Primitive Success Rates, Durations, and Calibrated Search Costs ($100$ Trials per Type)}
\label{tab:video-events}
\small
\setlength{\tabcolsep}{4pt}
\renewcommand{\arraystretch}{1.16}
\begin{tabular}{@{}lccc@{}}
\hline
\textbf{Primitive} & \textbf{Success (\%)} & \textbf{Median ($\mathrm{s}$)} & \textbf{Cost $\hat c(o)$ ($\mathrm{s}$)} \\
\hline
Unloaded step & $100$ & $15.0$ & $15$ \\
Loaded step & $100$ & $15.0$ & $15$ \\
$90^\circ$ turn & $100$ & $22.0$ & $20$ \\
$180^\circ$ turn & $100$ & $28.0$ & $30$ \\
Face unlocking & $96$ & $24.0$ & $25$ \\
Module unit pickup & $100$ & $9.0$ & $10$ \\
Module unit placement & $98$ & $8.0$ & $10$ \\
Face locking & $92$ & $27.0$ & $25$ \\
\hline
\end{tabular}
\end{table}

\textbf{1) Primitive duration analysis and cost calibration:} The eight primitive types defined in Section~\ref{sec:astar} were each tested $100$ times on the physical prototype. Success rates and median durations are reported in Table~\ref{tab:video-events}. Placement success was assessed within the three-attempt limit. The measured medians were discretized into time costs $\hat c(o)$ for the mid-level contact-face $A^*$ search, contributing to $g_{\mathrm{walk}}(s)$ in~\eqref{eq:walk-astar}.

Unlocking and locking failures mainly arose from deformation of the 3D-printed gears in the three-stage transmission after prolonged operation, which shifted the tightening-head pose. An integrated CNC-machined transmission is expected to reduce this drift. CMU placement occasionally jammed beside an existing CMU after prolonged operation because the two drive motors were rigidly synchronized. Adaptive differential control may reduce this binding.

\begin{figure}[!t]
\centering
\includegraphics[width=\columnwidth]{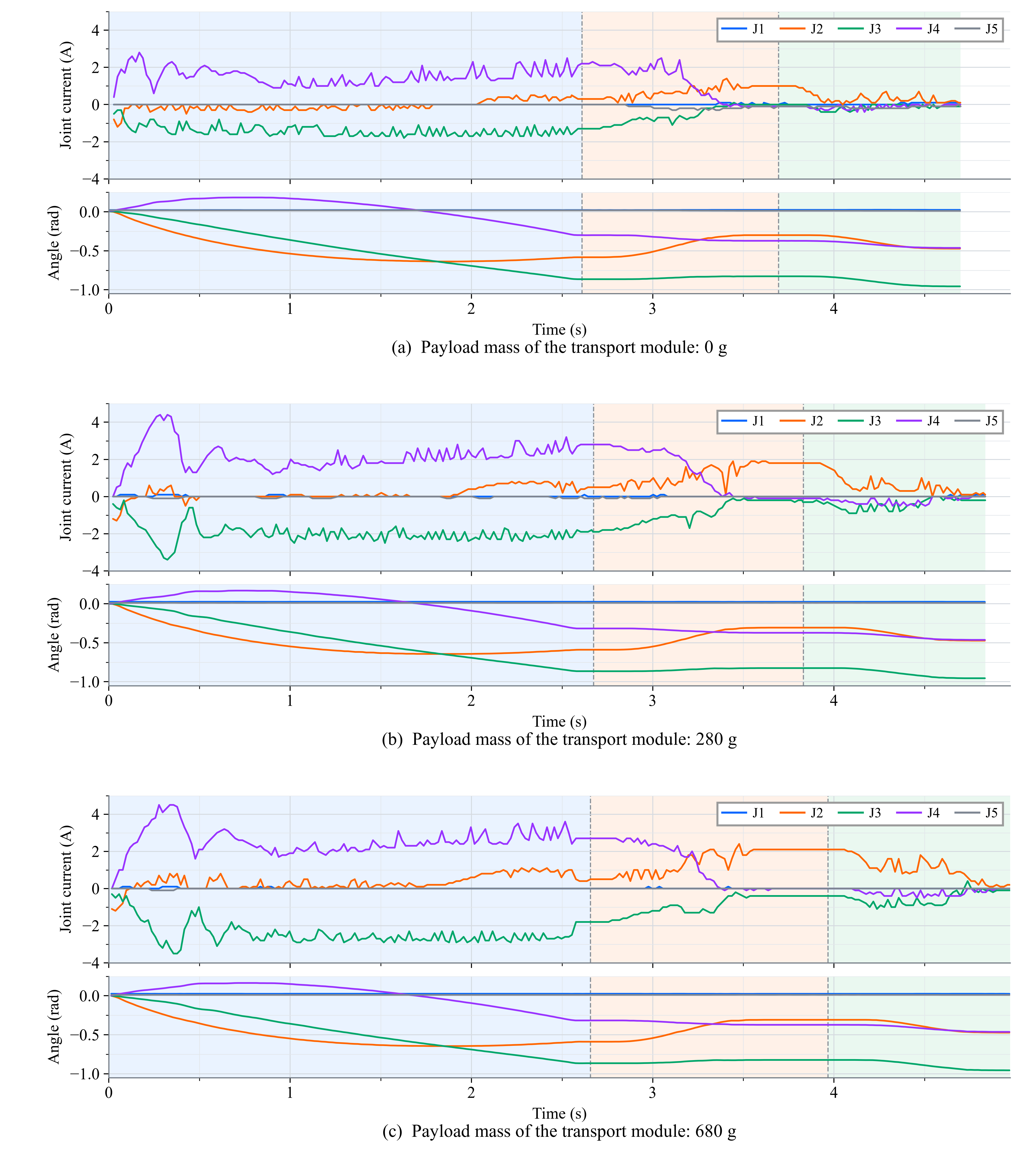}
\caption{Current and angle responses of locomotion-chain joints $J_1$--$J_5$ under payloads of (a) $0\,\mathrm{g}$, (b) $280\,\mathrm{g}$, and (c) $680\,\mathrm{g}$. Shaded regions denote the three operation phases.}
\label{fig:payload-current-angle}
\end{figure}

\textbf{2) Joint actuation and motion consistency under different payloads:} To evaluate how transported CMU payloads affect the actuation requirements and trajectory tracking of the locomotion chain ($J_1$--$J_5$), Fig.~\ref{fig:payload-current-angle} compares joint current and angle responses under payloads of $0\,\mathrm{g}$, $280\,\mathrm{g}$, and $680\,\mathrm{g}$. As the payload increases from $0\,\mathrm{g}$ to $680\,\mathrm{g}$, the peak and RMS currents of joints $J_3$ and $J_4$ rise monotonically during the initial lifting phase, whereas $J_2$ bears a larger share of the load during subsequent posture adjustments. In the first phase, the RMS currents of $J_3$ are approximately $1.45$, $2.06$, and $2.53\,\mathrm{A}$ and those of $J_4$ approximately $1.59$, $2.26$, and $2.64\,\mathrm{A}$ for the three payloads. Despite the increased torque demands, the joint angle trajectories and final angles remain consistent across all three payloads, and all three trials complete the operation.

\subsection{Transport, Assembly, and Reconfiguration}
\label{sec:transport-eval}

The primitives evaluated in Section~\ref{sec:primitive-eval} were combined into physical transport and assembly tasks in one and two dimensions. A planner-generated sequence was then executed to reconfigure an $11$-CMU lattice from an extended layout into a compact planar configuration. Large-scale execution was further examined in a $70$-CMU digital twin.

\begin{figure}[!t]
\centering
\includegraphics[width=\linewidth]{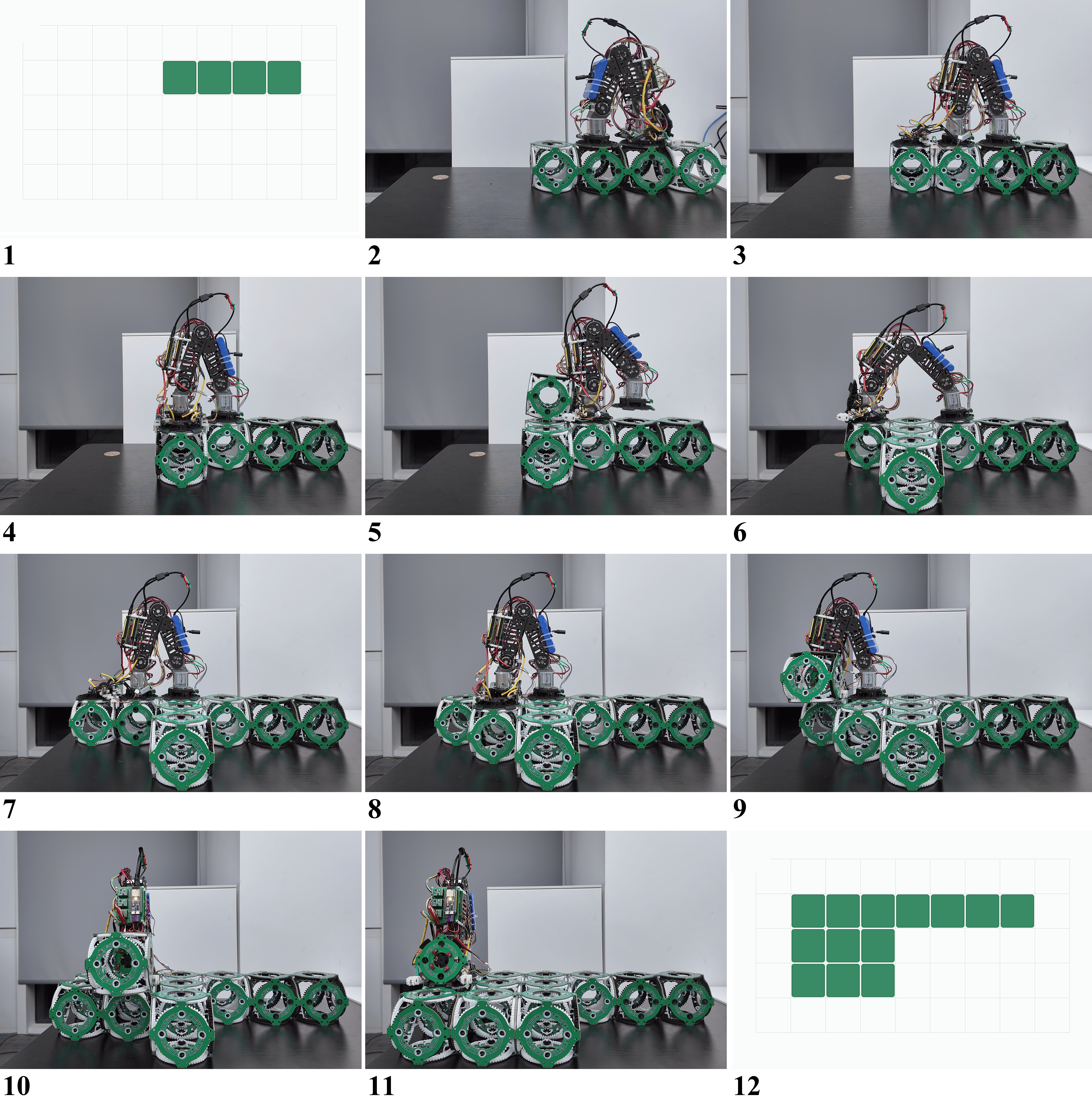}
\caption{Physical autonomous transport and planar assembly of nine CMUs: initial layout (1), keyframes (2--11), and final layout (12).}
\label{fig:transport}
\end{figure}

\textbf{1) One-dimensional linear extension and two-dimensional planar assembly:} In the one-dimensional task, the TR removed, transported, and docked $3$ CMUs from a fixed supply station to extend the lattice linearly, completing the task in approximately $300\,\mathrm{s}$. In the two-dimensional task (Fig.~\ref{fig:transport}), it transferred and assembled $9$ CMUs in approximately $1200\,\mathrm{s}$. Alternating stepping and same-face yaw enabled access to different target cells on the lattice surface. In both tasks, the TR traversed the assembled lattice in unloaded and loaded states and docked CMUs from multiple directions.

\begin{figure}[!t]
\centering
\includegraphics[width=\linewidth]{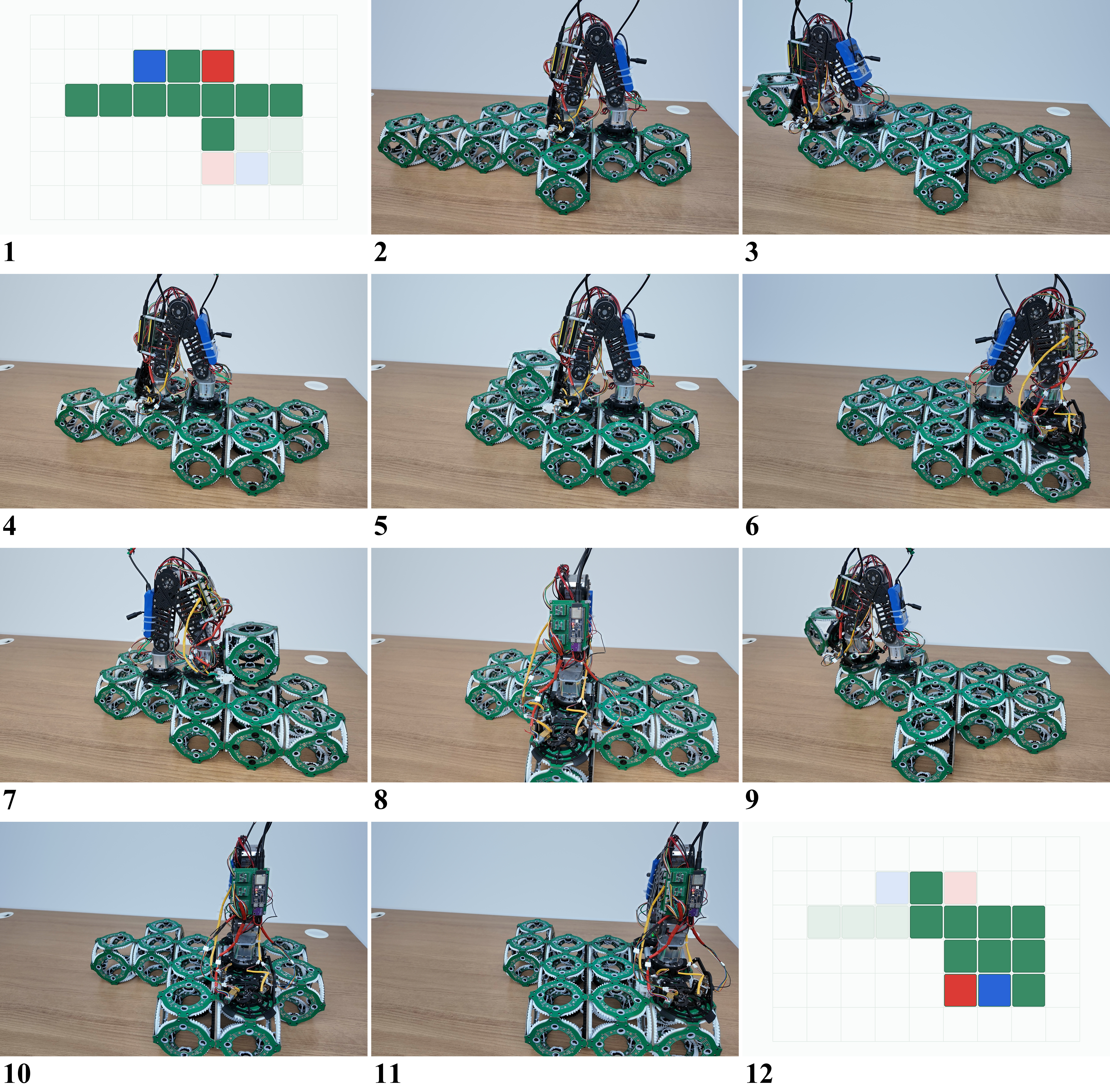}
\caption{Physical execution of a planner-generated reconfiguration sequence: initial configuration ($1$), intermediate actions ($2$--$11$), and final configuration ($12$). Red and blue denote two heterogeneous functional CMU types; faint cells show the counterpart configuration.}
\label{fig:physical-reconfiguration}
\par\vspace{10pt}
\includegraphics[width=\linewidth]{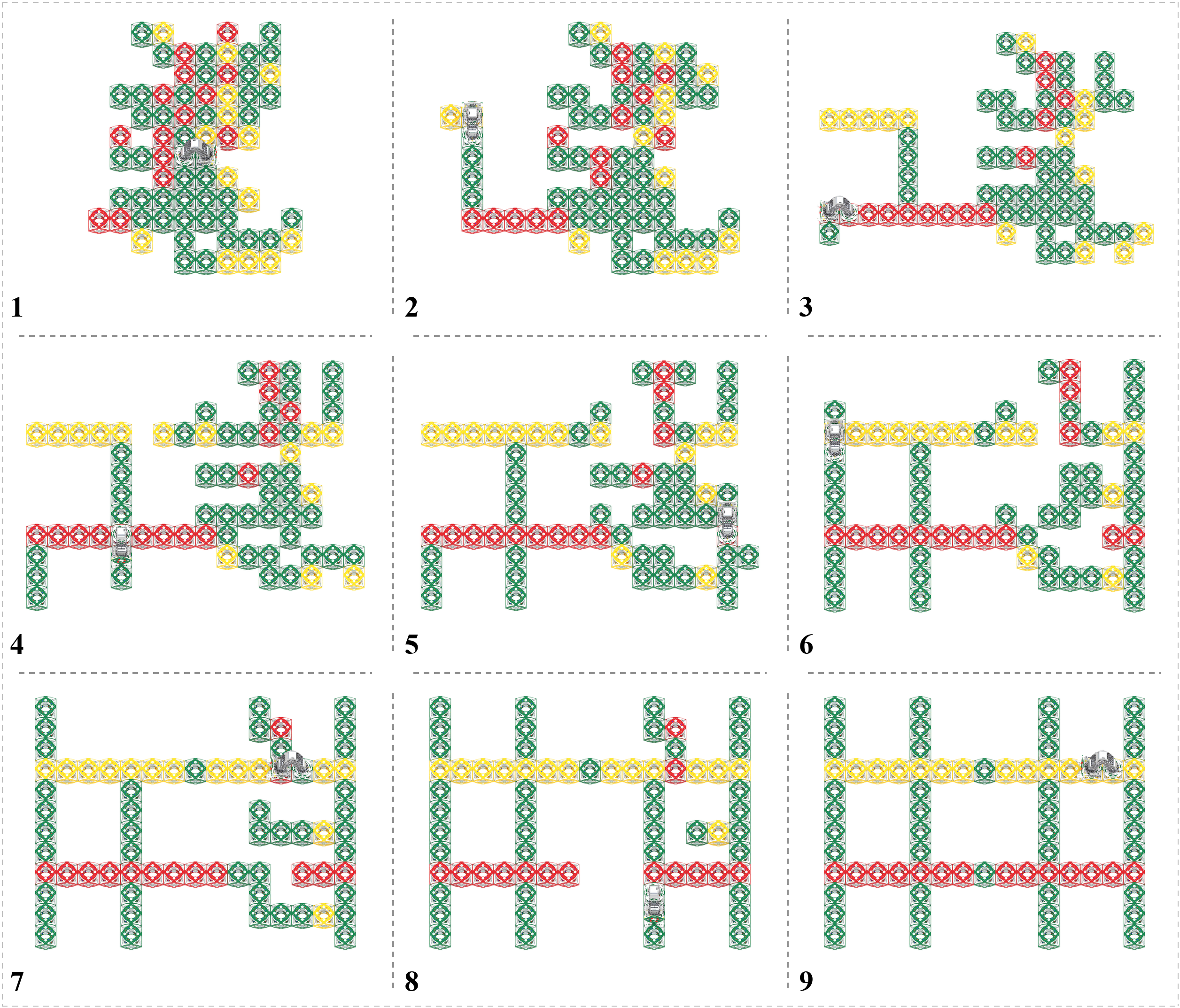}
\caption{Digital-twin reconfiguration of a $70$-module-unit lattice following the hierarchical plan (chronological order).}
\label{fig:transport-sim}
\end{figure}

\textbf{2) Planner-generated reconfiguration:} Fig.~\ref{fig:physical-reconfiguration} shows the physical execution of a planner-generated sequence that reconfigures an $11$-CMU lattice from an extended layout into a compact planar configuration. The sequence combines CMU removal, transport, and interface locking on the existing lattice. The physical assembly and reconfiguration experiments were each conducted three times, with all trials completed successfully. Full demonstrations are shown at accelerated playback speed in the supplementary video.

% Use independent column floats so body text can fill the space below each figure.

\textbf{3) Large-scale reconfiguration in a digital twin:} A $70$-module-unit lattice was reconfigured in a Blender digital twin incorporating the transport robot's kinematic constraints (Fig.~\ref{fig:transport-sim}). The macro-action sequence was expanded into contact-face primitives and continuous joint trajectories, maintaining structural connectivity and two-foot stance reachability throughout execution.

\subsection{Residual-Value Ablation, Search Efficiency, and Generalization to Larger Lattices}
\label{sec:planning-training}

\textbf{1) Training and evaluation protocol:} The planning network is trained via~\eqref{eq:joint-q-objective} and Algorithm~\ref{alg:value-learning} ($\lambda_{\mathrm E}=0.2$, $\lambda_{\mathrm U}=0.02$, $\vartheta=1$) on lattices of size $n\le100$ across a three-stage curriculum with $m_0=M(C_{\mathrm{ini}},G)\in[1,10]$, $[11,30]$, and $[31,100]$ initially misplaced CMUs. Using a single frozen checkpoint and identical legal action rules without wall-clock limits, we compare three planners: 1) \textbf{Proposed} (Algorithm~\ref{alg:value-search} with learned $Q_\theta$ in~\eqref{eq:residual-q-model}); 2) \textbf{Ablation} ($Q_{\mathrm{analytic}}(C,a,G)=1+M(C',G)$, omitting $r_\theta$); and 3) \textbf{Bidirectional} heuristic search~\cite{meng2026morphogenetic} (the expert planner basis in Section~\ref{sec:value-guided-search}). A case is solved if a verified macro-action sequence is found within budget $B$, which counts candidate edge pops for the two value-guided planners and expanded nodes for bidirectional search (granting the latter greater search effort at equal $B$, as each node expansion evaluates all incident edges). Three unseen test sets are evaluated: a benchmark split at $n=70$ ($176$ cases) and $n=100$ ($132$ cases); a dense set of $1{,}266$ cases over $n\in\{70,75,\dots,100\}$ ($150$--$213$ per size); and a scale-generalization set over $n\in\{110,120,\dots,150\}$ ($192$--$198$ per size, uniform $m_0$).

\begin{figure}[!t]
\centering
\includegraphics[width=\columnwidth]{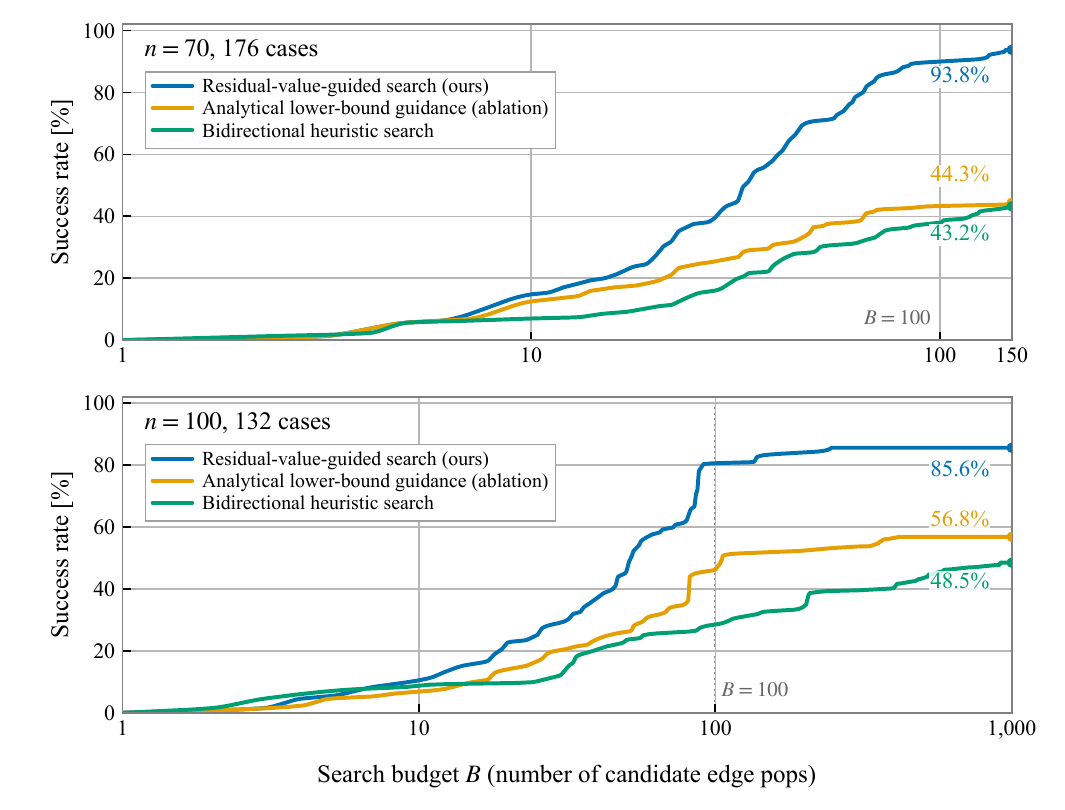}
\caption{Cumulative success rate versus search budget $B$ on the benchmark split: $n=70$ ($176$ cases, shown up to $B=150$) and $n=100$ ($132$ cases, $B\le1000$). Percentages indicate success at the displayed budget endpoints. Curves use shape-preserving monotone interpolation between measured budgets. The dotted line marks $B=100$. The two value-guided planners count candidate edge pops; bidirectional search counts expanded nodes.}
\label{fig:planning-evaluation}
\label{fig:planning-residual-ablation}
\end{figure}

\textbf{2) Residual ablation and search efficiency:} In Fig.~\ref{fig:planning-evaluation}, the proposed planner, ablation, and bidirectional search respectively solve $96.6\%$, $47.2\%$, and $73.3\%$ of cases at $n=70$ ($B=500$), and $85.6\%$, $56.8\%$, and $48.5\%$ at $n=100$ ($B=1000$). Most gains occur early: at $B=100$, the proposed method already reaches $89.8\%$ ($n=70$) and $80.3\%$ ($n=100$), versus $43.2\%$ and $46.2\%$ for the ablation and $37.5\%$ and $28.0\%$ for bidirectional search. Across the dense $1{,}266$-case set at $B=1000$, the proposed method solves $79.2\%$ ($1{,}003$ cases), outperforming the ablation ($28.0\%$, $354$ cases; $+51.2$ percentage points) and bidirectional search ($57.5\%$, $728$ cases), and prevailing at every lattice size ($70.3$--$93.5\%$ versus $48.8$--$77.6\%$ for bidirectional search). Even at $B=100$, its $73.3\%$ success rate ($928$ cases) exceeds both baselines at $B=1000$. Furthermore, increasing $B$ fails to rescue the ablation, which plateaus from $25.8\%$ ($B=100$) to $28.0\%$ ($B=500$ and $1000$) on the dense set and remains at $56.8\%$ ($B=500$ to $2000$) on the $n=100$ split. Because $M(C',G)$ cannot rank actions when connectivity and two-foot stance constraints require temporary detours, analytical search stalls on plateaus; the learned residual $r_\theta$ resolves this by estimating the extra relocation cost of such detours.

\begin{figure}[!t]
\centering
\includegraphics[width=\columnwidth]{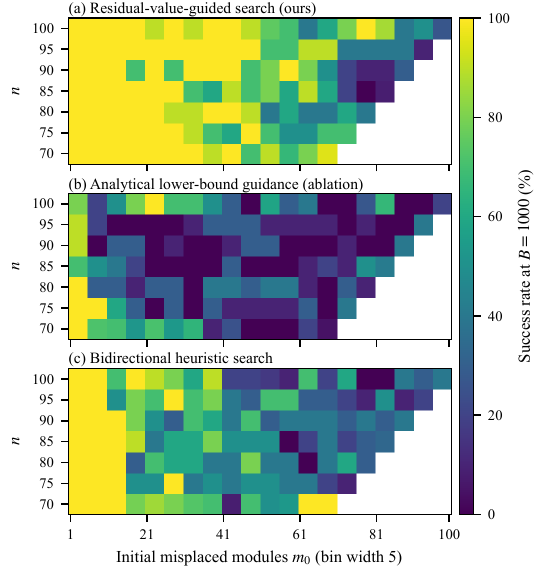}
\caption{Success rate at $B=1000$ on the dense $1{,}266$-case set, binned by CMU count $n$ and initial misplaced CMUs $m_0$ (bin width $5$). (a) Proposed method, (b) analytical lower-bound ablation, and (c) bidirectional search (budget in expanded nodes). White cells denote $m_0>n$ or bins without cases.}
\label{fig:planning-m0-heatmap}
\end{figure}

\textbf{3) Gains across initial misplacement:} Fig.~\ref{fig:planning-m0-heatmap} resolves success rates over $(n,m_0)$. For small misplacements ($m_0\le20$), the proposed, ablation, and bidirectional planners reach $99.0\%$, $51.7\%$, and $90.3\%$, respectively. The gap widens substantially as $m_0$ increases: the corresponding rates are $93.8\%$, $28.2\%$, and $67.5\%$ for $m_0\in[21,40]$; $79.2\%$, $18.3\%$, and $41.5\%$ for $m_0\in[41,60]$; and $58.4\%$, $13.4\%$, and $38.5\%$ for $m_0\in[61,80]$. The learned residual thus expands the solvable $(n,m_0)$ region most pronouncedly under moderate-to-large configuration differences.

\begin{figure}[!t]
\centering
\includegraphics[width=\columnwidth]{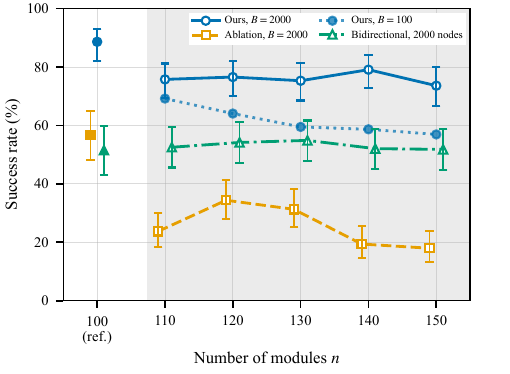}
\caption{Scale generalization to unseen lattice sizes $n=110$--$150$ ($B=2000$: hollow markers; $B=100$: dotted line; $n=100$ benchmark reference: filled markers; error bars: Wilson $95\%$ intervals).}
\label{fig:planning-scale-generalization}
\end{figure}

\textbf{4) Generalization to larger lattices and limitations:} On unseen sizes $n=110$--$150$ (Fig.~\ref{fig:planning-scale-generalization}; non-monotonic variations reflect independently generated sets per $n$), the proposed method at $B=2000$ solves $73.6$--$79.1\%$ of cases across sizes, compared with $17.9$--$34.4\%$ for the ablation and $51.8$--$54.9\%$ for bidirectional search ($73.6\%$ versus $17.9\%$ and $51.8\%$ at $n=150$). Even at $B=100$, the proposed method achieves $56.9$--$69.2\%$, outperforming both baselines at $B=2000$ across all $n$. At $n=150$ ($1.5\times$ the maximum training size), it solves all $78$ cases with $m_0\le60$ (versus $24$ and $61$ for the ablation and bidirectional baselines), though performance remains limited under near-complete rearrangements ($9/65$ cases solved for $m_0>100$).

\section{Conclusion}
\label{sec:conclusion}

CuBot-STT integrates CMU transport and reversible electromechanical fastening in one external TR, preserving internal payload space. Hierarchical planning connects learned configuration search with contact-face planning and constrained robot motion. Experiments verified $300\,\mathrm{N}$ normal and $490\,\mathrm{N}$ tangential capacities, power transmission across six CMUs, and autonomous $11$-CMU reconfiguration. A $70$-CMU digital twin demonstrated plan-to-motion expansion. Learned residual guidance outperformed analytical and bidirectional baselines, generalizing without fine-tuning from training sizes up to $100$ CMUs to $150$ CMUs.

Future work will integrate actuated, wheeled, and sensing CMUs for autonomous assembly and task-driven reconfiguration of functional modular robots.

% References follow the conclusion without a forced page break.
\bibliographystyle{\bibliographyprefix IEEEtran}
\bibliography{\bibliographyprefix refs}

% Author biographies follow the references on the same page.

\begin{IEEEbiography}[{\includegraphics[width=1in,height=1.25in,clip,keepaspectratio]{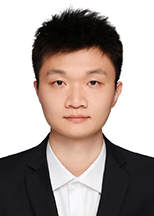}}]{Chongxi Meng}
received the B.E. degree in robotics engineering from Northeastern University, China, in 2021, and the M.Phil. degree in computer and information engineering from The Chinese University of Hong Kong, Shenzhen, China, in 2023. He is currently working toward the Ph.D. degree in intelligent science and technology with the Shanghai Research Institute for Intelligent Autonomous Systems, Tongji University, China. His research interests include reconfigurable robots.
\end{IEEEbiography}

\begin{IEEEbiography}[{\includegraphics[width=1in,height=1.25in,clip,keepaspectratio]{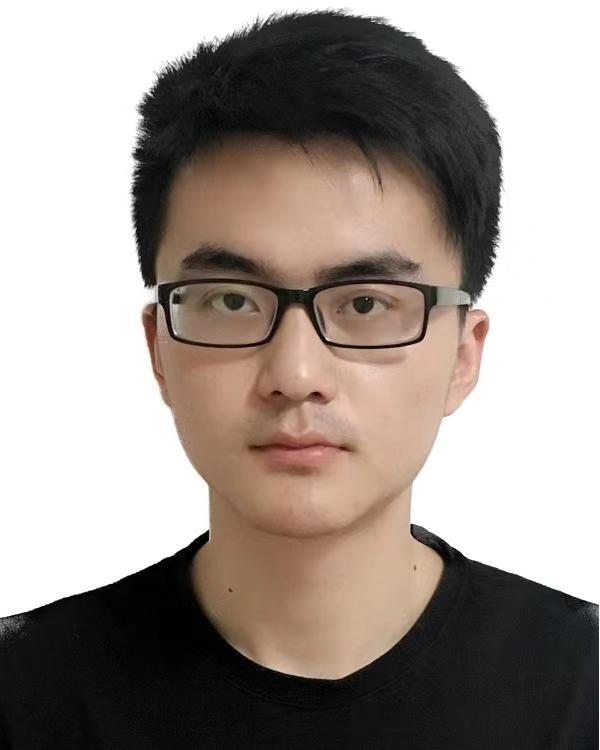}}]{Da Zhao}
received the bachelor's degree in mechanical engineering from China University of Mining and Technology, China, in 2019, and the M.Phil. degree in computer and information engineering from The Chinese University of Hong Kong, Shenzhen, China, in 2023. He is currently working toward the Ph.D. degree at The Chinese University of Hong Kong, Hong Kong. His current research interests include multiscale self-reconfigurable robots.
\end{IEEEbiography}

\begin{IEEEbiography}[{\includegraphics[width=1in,height=1.25in,clip,keepaspectratio]{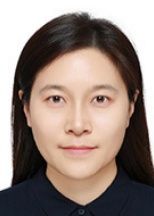}}]{Yanmin Zhou}
received the M.S. degree in control theory and control engineering from Tongji University, Shanghai, China, in 2011, and the Ph.D. degree in biomechanics from the University of Cambridge, Cambridge, U.K., in 2015. She is currently an Associate Professor with the Department of Control Science and Engineering, Tongji University. Her current research interests include bionics and the design and control of intelligent robots.
\end{IEEEbiography}

\begin{IEEEbiography}[{\includegraphics[width=1in,height=1.25in,clip,keepaspectratio]{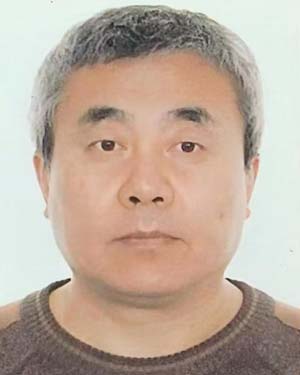}}]{Runjie Shen}
received the B.S. degree from Xi'an Polytechnic University, Xi'an, China, in 1995, the M.S. degree from Xi'an Jiaotong University, Xi'an, China, in 1998, and the Ph.D. degree from Zhejiang University, Hangzhou, China, in 2004. From 2004 to 2006, he was a Postdoctoral Researcher with Zhejiang University, where he became an Associate Researcher in 2006. Since 2011, he has been an Associate Researcher with Tongji University, Shanghai, China. His research interests include sensing technologies and nondestructive testing.
\end{IEEEbiography}

\begin{IEEEbiography}[{\includegraphics[width=1in,height=1.25in,clip,keepaspectratio]{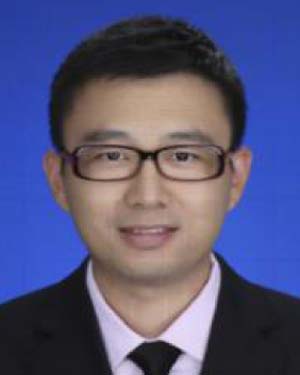}}]{Zhipeng Wang}
received the M.S. degree in mechanical manufacturing and automation from Zhejiang University, Hangzhou, China, in 2011, and the Ph.D. degree in control science and engineering from Tongji University, Shanghai, China, in 2015. He is currently an Associate Professor in the Department of Control Science and Engineering, Tongji University. His current research interests include biped robots, mechatronics, and dynamics.
\end{IEEEbiography}

\begin{IEEEbiography}[{\includegraphics[width=1in,height=1.25in,clip,keepaspectratio]{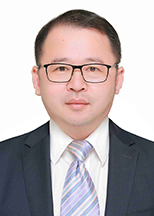}}]{Bin He}
(Senior Member, IEEE) received the B.S. degree in engineering machinery from Jilin University, Changchun, China, in 1996, and the Ph.D. degree in mechanical and electronic control engineering from Zhejiang University, Hangzhou, China, in 2001. He is currently a Professor with the Department of Control Science and Engineering, Tongji University, Shanghai, China. His current research interests include intelligent robot control, biomimetic microrobots, and wireless networks.
\end{IEEEbiography}

\end{document}